\documentclass{article}

\usepackage{PRIMEarxiv}

\usepackage[utf8]{inputenc} 
\usepackage[T1]{fontenc}    
\usepackage{hyperref}       
\usepackage{url}            
\usepackage{booktabs}       
\usepackage{amsfonts}       
\usepackage{nicefrac}       
\usepackage{microtype}      
\usepackage{lipsum}
\usepackage{fancyhdr}       
\usepackage{graphicx}       
\graphicspath{{media/}}     
\usepackage{natbib}
\usepackage{times}

\usepackage{amsmath,amsfonts,bm}

\def\eqref#1{equation~\ref{#1}}

\def\1{\bm{1}}

\DeclareMathAlphabet{\mathsfit}{\encodingdefault}{\sfdefault}{m}{sl}
\SetMathAlphabet{\mathsfit}{bold}{\encodingdefault}{\sfdefault}{bx}{n}

\usepackage{url}
\usepackage{graphicx}

\newtheorem{assumption}{Assumption}

\usepackage[capitalize,noabbrev]{cleveref}

\crefformat{equation}{Eq.~(#2#1#3)}
\Crefformat{equation}{Eq.~(#2#1#3)}
\Crefformat{Equation}{Eq.~(#2#1#3)}

\crefformat{section}{Sec.~#2#1#3}
\Crefformat{Section}{Sec.~#2#1#3}
\crefformat{figure}{Fig.~#2#1#3}
\Crefformat{Figure}{Fig.~#2#1#3}

\Crefformat{appendix}{App.~#2#1#3}
\Crefformat{Appendix}{App.~#2#1#3}
\Crefformat{app}{App.~#2#1#3}

\Crefformat{measure}{~#2#1#3}
\Crefformat{assumption}{~#2#1#3}
\Crefformat{definition}{~#2#1#3}

\newcommand{\LL}{\mathcal{L}}
\newcommand{\Erf}{\mathrm{Erf}}
\newcommand{\BB}{\mathcal{B}}

\newcommand{\Acc}{\mathcal{A}}
\newcommand{\trn}{\mathrm{tr}}
\newcommand{\gen}{\mathrm{gen}}

\newcommand{\tgen}{t^*_{\gen}}
\newcommand{\ttr}{t^*_{\trn}}
\newcommand{\rel}{\mathrm{rel}}
\newcommand{\abs}{\mathrm{abs}}

\title{Measuring Sharpness in Grokking}

\author{
  Jack Miller\thanks{Jack Miller and Patrick Gleeson contributed equally to this work. Contact email \textit{jack.miller@anu.edu.au}.}, Patrick Gleeson\footnotemark[1], Charles O'Neill, Thang Bui \\
  College of Engineering, Computing and Cybernetics \\
  The Australian National University \\
  Canberra\\
   \And
  Noam Levi \\
  School of Physics and Astronomy \\
  Tel-Aviv University \\
  Tel-Aviv, Israel \\
}

\begin{document}
\maketitle

\begin{abstract}
Neural networks sometimes exhibit \textit{grokking}, a phenomenon where perfect or near-perfect performance is achieved on a validation set well after the same performance has been obtained on the corresponding training set. In this workshop paper, we introduce a robust technique for measuring grokking, based on fitting an appropriate functional form. We then use this to investigate the sharpness of transitions in training and validation accuracy under two settings. The first setting is the theoretical framework developed by \cite{levi2023grokking} where closed form expressions are readily accessible. The second setting is a two-layer MLP trained to predict the parity of bits, with grokking induced by the concealment strategy of \cite{miller2023grokking}. We find that trends between relative grokking gap and grokking sharpness are similar in both settings when using absolute and relative measures of sharpness. Reflecting on this, we make progress toward explaining some trends and identify the need for further study to untangle the various mechanisms which influence the sharpness of grokking.
\end{abstract}

\section{Introduction}

\addtocounter{footnote}{-1}

After the recent discovery of grokking in transformers completing algorithmic tasks \citep{power-grokking:2022}, many have worked to better understand the phenomenon \citep{parity-prediction:2023, rulesbased2022grokking,davies2023unifying, levi2023grokking, liu2023omnigrok, lyu2023dichotomy, miller2023grokking, nanda2023progress, varma-grokking-circuits:2023}. Grokking has now been demonstrated in multilayer perceptrons (MLPs) \citep{merrill2023tale}, Gaussian processes (GPs) \citep{miller2023grokking} and linear regression \citep{miller2023grokking, levi2023grokking}, as well as on non-algorithmic tasks \citep{liu2023omnigrok}. Recent theoretical work has also shown that grokking can be provably induced in linear student-teacher models \citep{levi2023grokking} and homogeneous neural networks \citep{lyu2023dichotomy}.

While considerable attention has been devoted to the delay between training and test performance in grokking, little has been done to examine the `sharpness' (suddenness) of transitions.\footnote{The only recent mention we could find was in Appendix B of \cite{lyu2023dichotomy}.} We believe this is an essential feature of grokking and worthy of further study. To examine sharpness, we leverage the fact that jumps in training and validation accuracy can often be approximated by the same functional form. By fitting this function, we obtain mutually consistent measures of time-to-performance and sharpness. This gives a principled way to study these variables across a range of settings.

\section{Measuring Grokking}
To compare properties of the validation accuracy in grokking across tasks and architectures, we are interested in observing variables which are defined \emph{relative to the training accuracy}, as this avoids sensitivity to the overall timescale or difficulty of learning. As we will see, such comparisons are particularly principled when the training accuracy has a similar S-shaped form to the test accuracy, which (we observe) is typically the case whenever the initial training jump is not immediate. 

\begin{assumption}[Shared Functional Form]
    \label{assumption:shared-functional-form}
    We assume that both $\Acc_{\trn}$ and $\Acc_{\gen}$ are well-approximated by the same (S-shaped) functional form $f(x)$ up to scaling and translation
\begin{equation}\label{eqn:functional-form}
    \Acc_{\trn/\gen}(t) = a_{\trn/\gen}f\left(s_{\trn/\gen}
    [t-t^*_{\trn/\gen}] + \theta\right) + b_\mathrm{tr/gen},
    \end{equation}
    where $\mathcal{A}_\mathrm{tr/gen}$ is the training or validation accuracy, and $t$ is epoch. Here all parameters are scalars depending on the choice of $f$; for typical $f$ we have $a,s,t^*\in\mathbb{R}^+$ and $\theta,b\in \mathbb{R}$.
\end{assumption}

Generally, parameters $a$ and $b$ are derived from the total accuracy jump and the baseline (i.e. random-chance) accuracy respectively, though their exact interpretation depends on the vertical range of $f$. Even after constraining these, we observe empirically that, for our purposes, Assumption \ref{assumption:shared-functional-form} holds in Levi's linear estimator model and in our MLP task.\footnote{It also looks to be the case in the MNIST experiments of \cite{liu2023omnigrok} and non-MLP examples of \cite{miller2023grokking}.}  Further, variables $s$ and $t^*$ serve as natural measures of {\it sharpness} and {\it jump-time} respectively.

We can see in \cref{eqn:erf-form} that $t^*$ represents the occurrence-time of the jump, as measured by the particular accuracy
\[\Acc(t^*) = af(\theta) + b =: \Acc_0.\]
Notably, $\Acc_0$ is \emph{independent of $s$, $t^*$ and whether we consider $\Acc_\trn$ or $\Acc_\gen$}.\footnote{if $a$ and $b$ are shared which they are in the traditional cases of grokking.} Thus defining jump time by $t^*$ naturally reflects the traditional use of an accuracy threshold to define grokking, where this threshold can be modified by choosing an appropriate $\theta$. The use of $s$ to quantify sharpness is similarly principled since this variable controls the horizontal compression of the jump curve about $t^*$, and has a fixed proportionality to the accuracy gradient $\Acc'(t^*)$ there
\[\Acc'(t^*) = asf'(\theta).\]

Under Assumption \ref{assumption:shared-functional-form}, we also obtain simple expressions for the \emph{relative grokking gap} ($m$) and \emph{relative sharpness} ($R_\rel$) solely in terms of $t^*$ and $s$ respectively
\begin{equation}
    \label{eqn:relative-grokking-gap-definition}
    m = \frac{t^*_\mathrm{gen} - t^*_\mathrm{tr}}{t^*_\mathrm{tr}} = \frac{t^*_\mathrm{gen}}{t^*_\mathrm{tr}} - 1 ,
\end{equation}
\begin{align}
    \label{eqn:relative-accuracy-gradient}
    R_{\rel} = \frac{\Acc_{\gen}'(\tgen)}{\Acc'_{\trn}(\ttr)} = \frac{s_{\gen}}{s_{\trn}}.
\end{align}

Computing the quantities introduced above via curve-fitting has two advantages: (1) robustness to noise and (2) providing a natural check of the underlying assumptions (via the fitting quality). In our settings we choose $f(x)=\mathrm{Erf}(x)$ as a generic S-shaped curve. For general S-shapes, the fit is best in the linear region of the jump; we therefore take $\theta=0$, corresponding with an accuracy threshold at the midpoint. This also represents the maximum accuracy gradient, which we believe is the most intuitive measure of jump sharpness (given an S-shaped curve).

\begin{assumption}[Error Function Fitting]
    \label{assumption:error-function-fitting}
    We assume that both $\mathcal{A}_\mathrm{tr}$ and $\mathcal{A}_\mathrm{gen}$ are well approximated by the functional form
    \begin{equation}
        \label{eqn:erf-form}
        \mathcal{A}_\mathrm{tr/gen}(t) = a \mathrm{Erf}\left(s_\mathrm{tr/gen}(t-t^*_\mathrm{tr/gen})\right) + b , 
    \end{equation}
    where $a$ and $b$ are determined to achieve the range $[c,d]$ of model accuracy values. Namely, $a = d-c$ and $b = \frac{d-c}{2}$.
\end{assumption}

In this workshop paper, we study two measures of sharpness. \textbf{Measure 1} is the relative sharpness $R_{\mathrm{rel}}$ as defined previously. Coincidentally, because the maximum gradient of $\Erf$ occurs at its midpoint, in this case $R_{\rel}$ can also be expressed as
\begin{equation}
    R_{\mathrm{rel}} = \frac{\max_{t} \Acc_{\gen}'(t)}{\max_{t} \Acc'_{\trn}(t)}.
\end{equation}
Our second measure (\textbf{Measure 2}) is the \emph{absolute} gradient
\begin{align}
    R_{\mathrm{abs}} = \Acc_{\gen}'(t^*) = \frac{2as_\mathrm{gen}}{\sqrt{\pi}}.
\end{align}
which is equivalent to the maximum accuracy gradient.

\section{Experiments}

We report trends between $m$ and our sharpness measures in two settings. The first setting is the linear student teacher setup of \cite{levi2023grokking} which we investigate in \cref{sec:linear-student-teacher}. The second setting is the concealed parity prediction problem introduced by \cite{miller2023grokking} which is dealt with in \cref{sec:mlp}.

\subsection{Linear Student Teacher Setup}
\label{sec:linear-student-teacher}

\cite{levi2023grokking} considers linear student and teacher models in which the student is trained via gradient descent with a mean squared error (MSE) loss to reproduce the output of the teacher model under $N$ samples of i.i.d. data drawn from $\mathcal{N}\left(0, I_{d_{\mathrm{in}} \times d_{\mathrm{in}}}\right)$.\footnote{Since the framework has been described previously in the literature, we only provide a brief summary and refer the reader to the original paper for greater detail.} In the framework, we take $d_{in}, N \to \infty$ with $\lambda := d_{in}/N$ fixed. An example is considered correctly classified if its square-loss falls below a threshold $\epsilon$; the overall model accuracy is then given by
\begin{align}
    \label{eqn:accuracy-in-noam}
    \mathcal{B}_{\trn/\gen} = \mathrm{Erf}\left(\sqrt{\frac{\epsilon}{2 \mathcal{L}_{\trn/\gen}}}\right).
\end{align}
In this workshop paper we take $\epsilon =10^{-10}$; this ensures that training performance is delayed and hence exhibits a similar S-shape to the generalisation accuracy. It also ensures that both jumps occur in the long-time limit (below).

In the gradient flow limit (learning rate $\eta = \eta_0 dt$ with $dt \rightarrow 0$, $\eta_0\in\mathbb{R}^+$), the exact dynamics of the training and validation loss functions are given by
\begin{align}
    \label{eqn:training-loss-over-time}
    \mathcal{L}_\mathrm{tr}(t) = D_0^T e^{-4 \eta_0 \Sigma_\mathrm{tr} t} \Sigma_\mathrm{tr} D_0, \hspace{1em} \mathcal{L}_\mathrm{gen} = D_0^T e^{-4 \eta_0 \Sigma_\mathrm{tr} t} D_0,
\end{align}
where $S, T\in \mathbb{R}^{d_\mathrm{in}}$ are the student and teacher weights respectively, $D := S-T$ with $D_0 := D(t=0)$, and $\Sigma_{\mathrm{tr}}\in\mathbb{R}^{d_\mathrm{in}\times d_\mathrm{in}}$ denotes the training-set data covariance. 

Test-loss dynamics are given exactly in terms of the training loss by
\begin{equation}\label{eqn:dLgen}
    \frac{d\LL_{\gen}}{dt} = -4\eta_0 \LL_{\trn}.
\end{equation}
Further, the expected training loss permits a good approximation when $\eta_0 t \gg \sqrt{\lambda}$ which is reported in \cref{eqn:closed-form-training-loss-over-time}. This approximation not only allows computation of $\BB_{\trn}(t)$ but also of $\BB_{\gen}(t)$ via integration of \cref{eqn:dLgen}.
\begin{align}
    \label{eqn:closed-form-training-loss-over-time}
    \mathcal{L}_\mathrm{tr}(t)
    \approx \frac{e^{-4 \eta_0 (1-\sqrt{\lambda})^2 t}}{16 \sqrt{\pi} \lambda^{3/4} (\eta_0 t)^{3/2}}.
\end{align} 

Using the expressions derived for the training and generalisation accuracies, we examined trends in our sharpness measures by varying $\lambda$ toward $1$ thus increasing $m$. Then for each set of training curves, we fit the form of \cref{eqn:erf-form} using $a = \frac{1}{2}$ and $b = \frac{1}{2}$. The result of this can be see in \cref{fig:linear-measures}. One can also see the empirical fits of \cref{eqn:erf-form} in \cref{appendix:empirical-evidence-for-function-fits} and the trends in log-log space in \cref{appendix:log-log-plots}.

\begin{figure}[htb]
    \centering
    \includegraphics[width=.9\textwidth]{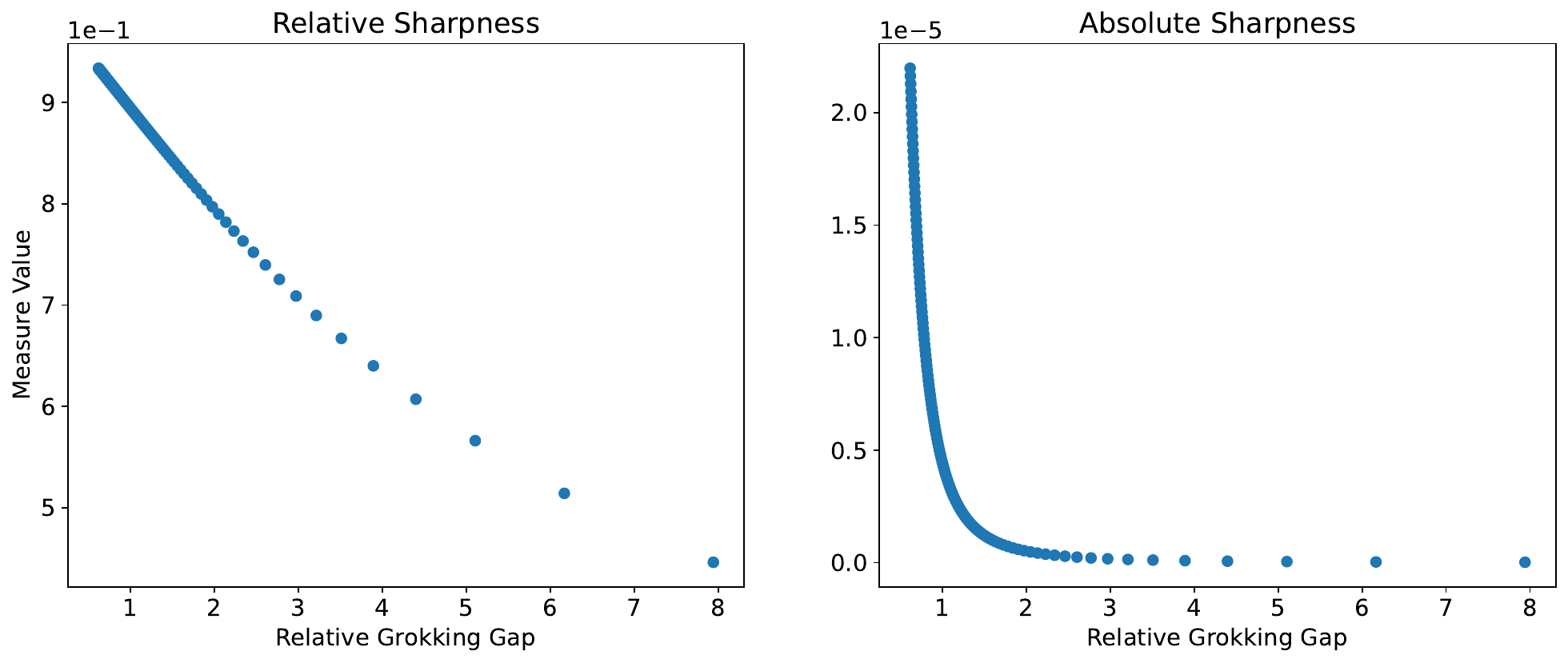}
    \caption{Relationship between relative grokking gap and both measures under the linear framework. For both measures we took $\lambda \in [1.003, 1.1]$, $\epsilon = 10^{-10}$ and $\eta_0 = 0.01$. Note that lines simply connect the data points and do not represent a trend. See \cref{appendix:log-log-plots} for linear trends in log-log space.}
    \label{fig:linear-measures}
\end{figure}

\subsection{MLP with Concealed Parity Prediction}
\label{sec:mlp}

Recently, \cite{miller2023grokking} showed that one can control the level of grokking (and thus the relative grokking gap) on algorithmic datasets with a 2-layer MLP. In particular, they found that adding spurious dimensions drawn i.i.d. from a $\mathrm{Ber}(1/2)$ distribution over $X \in \left\{0,1\right\}$ (i.e. random bits) to input examples reliably increases the relative grokking gap. We use this technique to empirically explore the value of $R_\mathrm{rel}$ and $R_\mathrm{abs}$ by varying the number of spurious dimensions on the parity prediction problem. As with the linear case we fit \cref{eqn:erf-form}. Unlike the linear case however, the baseline (random-chance) accuracy for parity prediction is 50\% and we accordingly set $a = \frac{1}{4}$ and $b = \frac{3}{4}$. The result is presented in \cref{fig:empirical-on-mlp}. As in the previous section, empirical fits and log-log plots can be seen in \cref{appendix:empirical-evidence-for-function-fits} and \cref{appendix:log-log-plots}.

\begin{figure}[htb]
    \centering
    \includegraphics[width=1\textwidth]{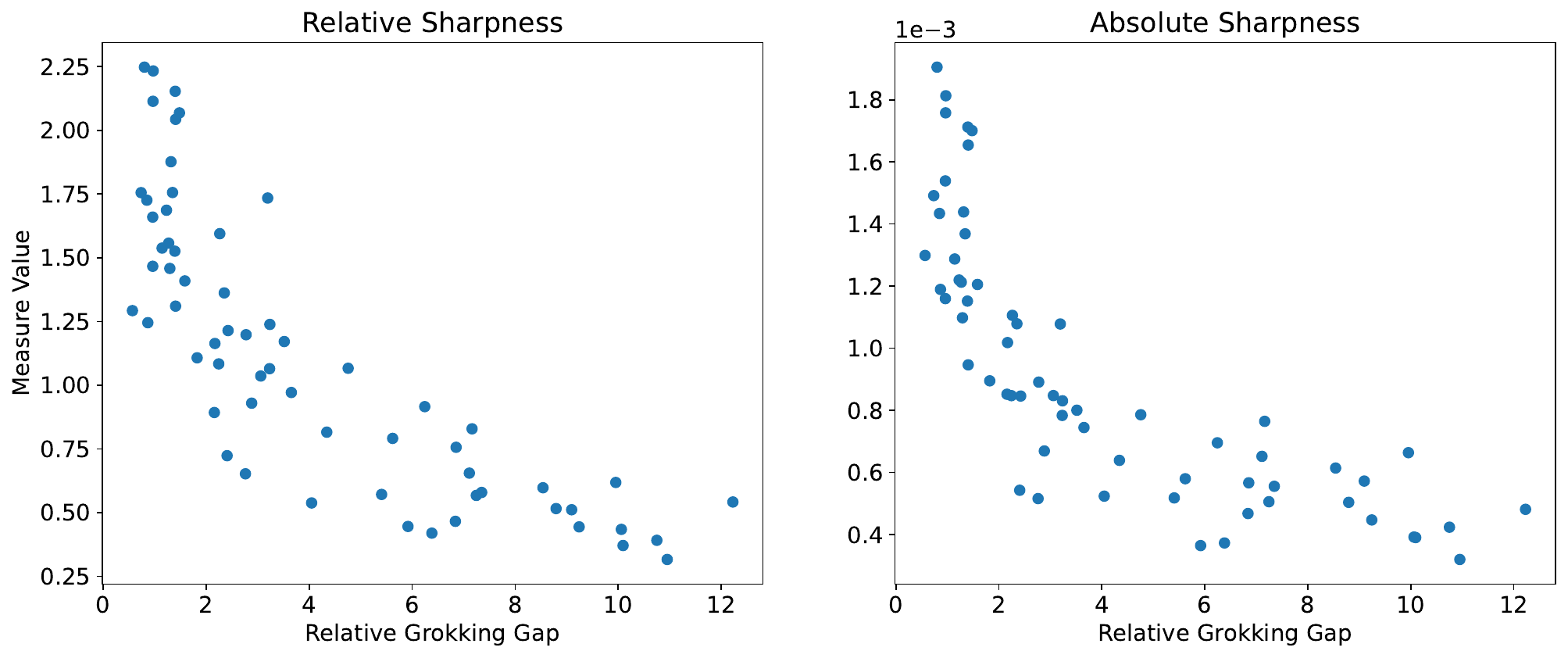}
    \caption{Relationship between relative grokking gap and $R_\mathrm{rel}$ and $R_\mathrm{abs}$. Each dot represents a training run.}
    \label{fig:empirical-on-mlp}
\end{figure}

\section{Discussion}
In this short paper, we first introduced a means of quantifying relative measures of grokking via fitting the same S-shaped functional form to both train- and test-accuracy curves. This gives rise to natural and empirically-robust measures of sharpness and jump-occurrence time, which reflect a given choice of accuracy threshold as in previous works. In our case, we choose the $\Erf$ function and accordingly measure at the halfway-accuracy point, as this represents both the sharpest point and the point of best fit. We then proposed two reasonable measures $R_\rel$ and $R_\abs$ to investigate sharpness. In experimental settings we found that both appeared to decrease with relative grokking gap. We now briefly discuss potential explanations, before concluding with directions for further work.

One might be able to explain the sharpness decrease in the case of the MLP by applying the grokking mechanism hypothesised by \cite{miller2023grokking}. One would begin by noting that solution search is guided by data fit and complexity. Further, one would assume that the effect of the complexity penalty is slower than that of data fit. In the case of low numbers of spurious dimensions, initial solutions may exist in regions without significant additional complexity relative to the task at hand. As such, when one reaches the appropriate complexity for the task, data fit is still shaping the solution leading to a sharper transition in validation accuracy. Alternatively, for larger spurious dimension counts (and thus larger $m$), general solutions will form later during optimisation. In these cases, it may be the slower complexity penalty which dominates changes in accuracy, thus resulting in a softer transition. Interestingly, if this logic were true, it could not be applied to the linear case which does not include regularisation.

In the linear setting, relationships $m(\lambda)$, $R_{\rel/\abs}(\lambda)$ and hence $R_{\rel/\abs}(m)$ can be approximated analytically. This may prove insightful; but for now we give a rough sketch as to why we see the trend we do in absolute measure by appealing to the chain rule. Indeed, we know that
\begin{equation}
    \partial_{t} \mathcal{A}(\mathcal{L}(t)) = \partial_t \mathcal{L}(t) \cdot \partial_\mathcal{L} \mathcal{A}(\mathcal{L}).
\end{equation}
We also know $\mathcal{L}(t)$ is of approximately exponential type ($e^{-t} / t^{3/2}$) for the long time limit and that $\partial_\mathcal{L} \mathcal{A}(\mathcal{L})$ is approximately the same for training and generalisation at the point of grokking. Thus, as one increases the relative grokking gap via $\lambda$ (consequently progressing along $\mathcal{L}(t)$), we should get an approximately exponential decrease in the gradient of validation accuracy since the derivative of $\mathcal{L}(t)$ decays approximately exponentially.

Despite providing these initial explanations for trends, there is more work to be done to validate them empirically in the case of the MLP or theoretically in the case of the linear estimators. In addition to this, there are yet further limitations to address and threads to investigate, which might further narrow the gap between theory and practice in this subfield of deep learning.
\begin{enumerate}  
    \item We claimed that delayed accuracy jumps are generally S-shaped, \emph{both for training and test curves}. Does this hold across existing works? Universality of a reasonably-sharp S-shape might suggest a thresholding mechanism akin to \cite{levi2023grokking}. Our framework also needs extension to cases where the training jump is immediate.\footnote{At present, our choice of 50\% accuracy threshold makes the fitting work in this case (e.g. our MLP)}
    \item We believe the parameters used to induce grokking in the linear-estimator case ($\lambda$) and MLP investigation (concealment) can be unified under the same framework.
    \item Can other interesting properties be extracted by fitting grokking curves, potentially using more exotic functional forms?
    \item Can this approach be used to identify (in)consistencies across existing works under a widely-applicable measure of grokking? We note that the framework generalises naturally to cases where final validation accuracy is not perfect.
\end{enumerate}
As a group we are continuing to work on these questions and limitations. We hope that this short workshop paper has demonstrated the utility of fitting error functions to grokking data and that the trends we have uncovered might inspire further theoretical advances to explain them. 

\section*{Supplementary Material}

Relevant code for this workshop paper can be found in the branch \texttt{feat/sharpness-and-gap} of the repository \href{https://github.com/jackmiller2003/tiny-gen}{tiny-gen}.  The script for generating the results in the linear estimators case is \texttt{scripts/sharpness-in-linear-estimators.ipynb}. Alternatively, the relevant script for the MLP case is contained within \texttt{tiny-gen.py} under the function name \texttt{experiment\_sharpness\_and\_grokking\_gap}.

\newpage

\bibliography{templateArxiv}
\bibliographystyle{templateArxiv}

\newpage

\appendix

\section{Empirical Evidence for Error Function Fits}
\label{appendix:empirical-evidence-for-function-fits}

In this section, we provide two figures which demonstrate the fit of error functions to the grokking data we use. Note how well \cref{eqn:erf-form} seems to work near the midpoint which we use to measure sharpness and the grokking gap.

\begin{figure}[ht!]
    \centering
    \includegraphics[width=1\textwidth]{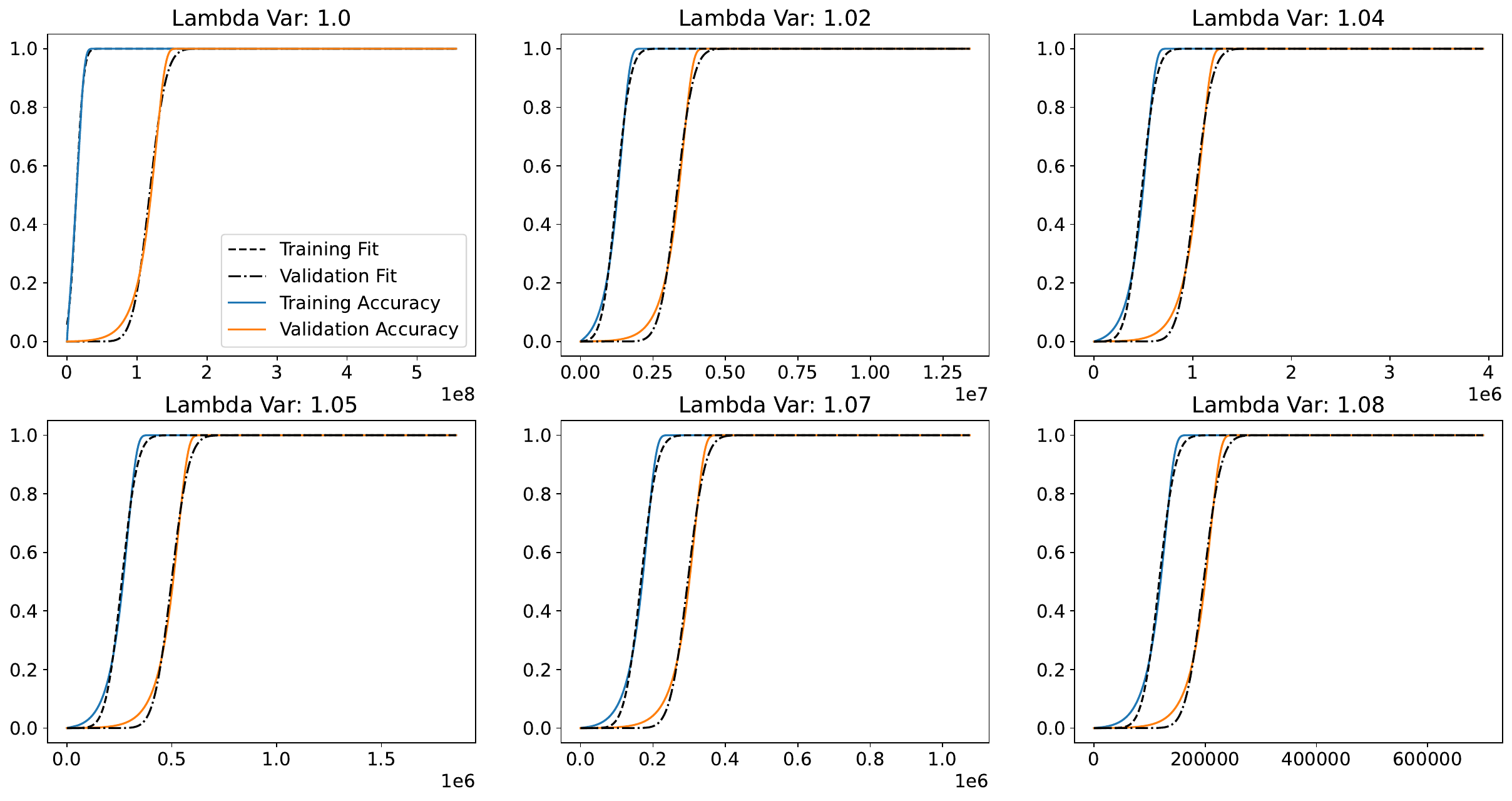}
    \caption{Visual evidence of \cref{eqn:erf-form} fit in the linear estimators setting. Note for numerical reasons, the domain for each $\lambda$ is distinct.}
    \label{fig:justification-for-erf-in-linear}
\end{figure}

\begin{figure}[ht!]
    \centering
    \includegraphics[width=1\textwidth]{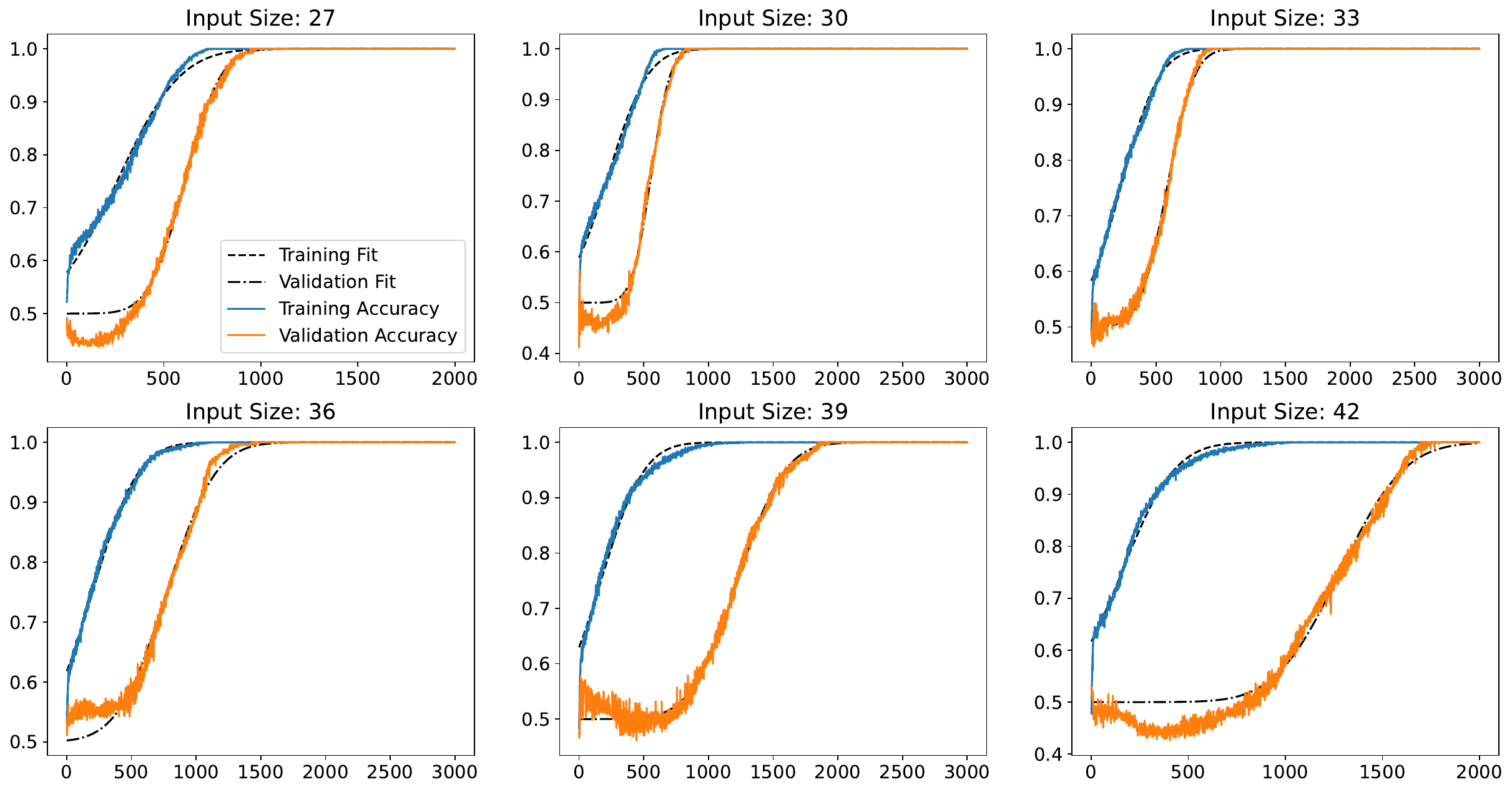}
    \caption{Visual evidence of \cref{eqn:erf-form} fit in the MLP setting. Note that input size refers to the changing size of examples resulting from the addition of further spurious dimensions.}
    \label{fig:justification-for-erf-in-mlp}
\end{figure}

\newpage

\section{Log-Log Plots for Experiments}
\label{appendix:log-log-plots}

Below we present the plots for the experiments on log-log scales. Additionally, we have included a linear fit in log-log space, the coefficients of which we report in the figures.

\begin{figure}[ht!]
    \centering
    \includegraphics[width=1\textwidth]{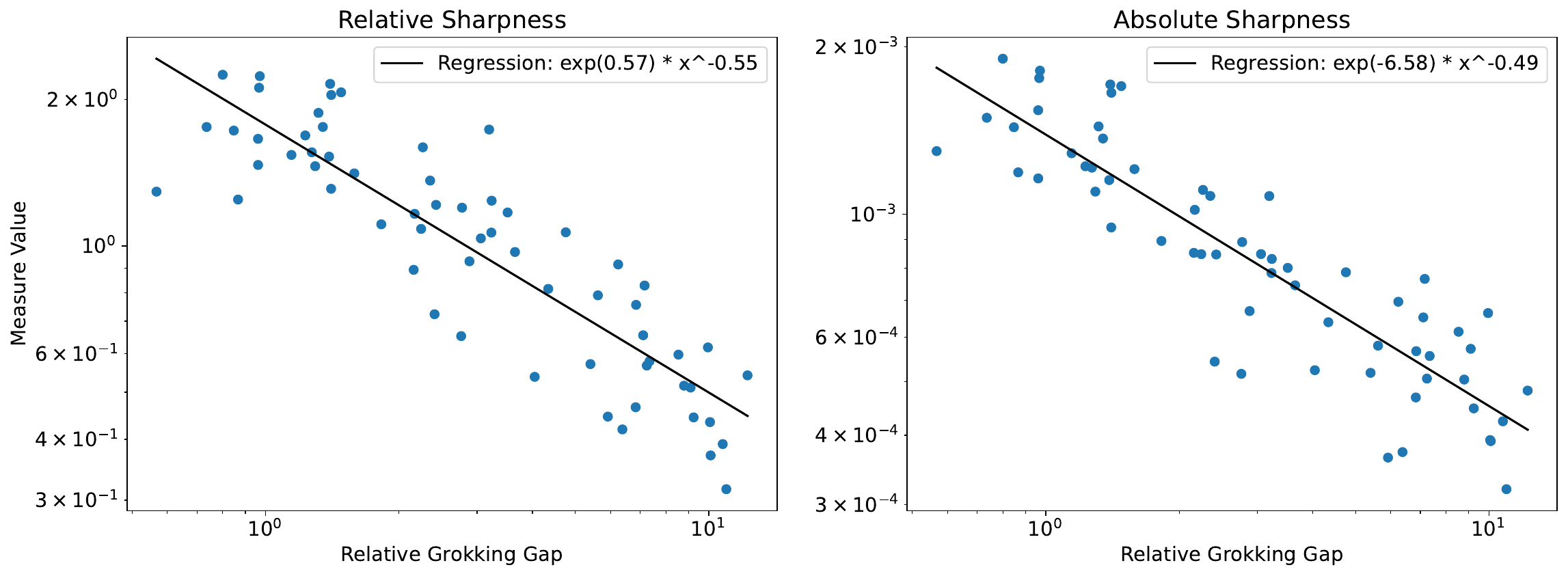}
    \caption{Same setting as in \cref{fig:empirical-on-mlp} but with a log-log scale and linear trend.}
    \label{fig:empirical-on-mlp-log-log}
\end{figure}

\begin{figure}[ht!]
    \centering
    \includegraphics[width=1\textwidth]{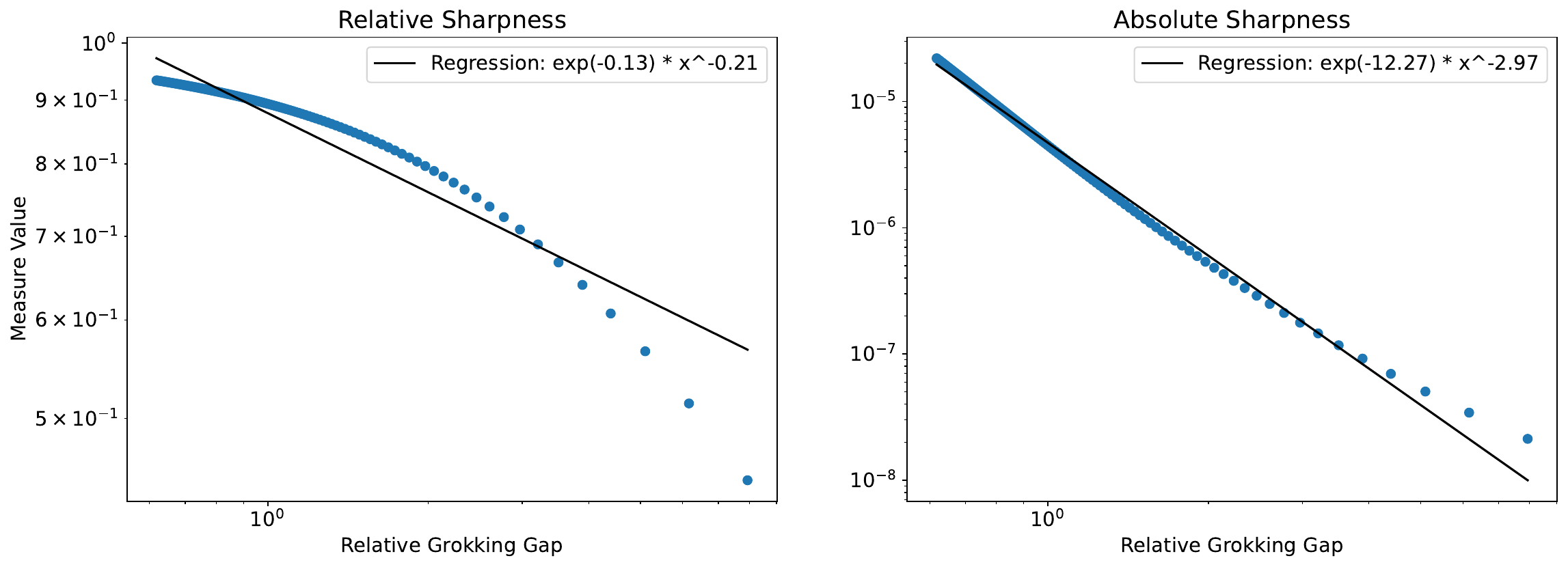}
    \caption{Same setting as in \cref{fig:linear-measures} but with a log-log scale and linear trend.}
    \label{fig:linear-measure-log-log}
\end{figure}

\end{document}